\documentclass[sigconf, screen]{acmart}

\graphicspath{ {./media/} }
\setcopyright{acmcopyright}

\AtBeginDocument{%
  \providecommand\BibTeX{{%
    \normalfont B\kern-0.5em{\scshape i\kern-0.25em b}\kern-0.8em\TeX}}}

\copyrightyear{2019}
\acmYear{2019}
\acmConference[MMSports '19]{2nd International Workshop on Multimedia Content Analysis in Sports}{October 25, 2019}{Nice, France}
\acmBooktitle{2nd International Workshop on Multimedia Content Analysis in Sports (MMSports '19), October 25, 2019, Nice, France}
\acmPrice{15.00}
\acmDOI{10.1145/3347318.3355517}
\acmISBN{978-1-4503-6911-4/19/10}

\begin{document}
\fancyhead{}

\title[CNN-based Segmentation Architecture for Ball Detection]
{Real-time CNN-based Segmentation Architecture for Ball Detection in a Single View Setup}

\newcommand{\tbr}[1]{{#1}}

\author{Gabriel Van Zandycke}
\email{gabriel.vanzandycke@uclouvain.be}
\orcid{0000-0002-8384-4166}
\affiliation{%
    \institution{UCLouvain}
    \city{Louvain-la-Neuve}
    \country{Belgium}
}

\author{Christophe De Vleeschouwer}
\email{christophe.devleeschouwer@uclouvain.be}
\orcid{0000-0001-5049-2929}
\affiliation{%
    \institution{UCLouvain}
    \city{Louvain-la-Neuve}
    \country{Belgium}
}

\begin{abstract}
This paper considers the task of detecting the ball from a single viewpoint in the challenging but common case where the ball interacts frequently with players while being poorly contrasted with respect to the background.
We propose a novel approach by formulating the problem as a segmentation task solved by an efficient CNN architecture. To take advantage of the ball dynamics, the network is fed with a pair of consecutive images.
Our inference model can run in real time without the delay induced by a temporal analysis. We also show that test-time data augmentation allows for a significant increase the detection accuracy. As an additional contribution, we publicly release the dataset on which this work is based.
\end{abstract}

\ccsdesc[500]{Computing methodologies~Interest point and salient region detections}
\ccsdesc[500]{Computing methodologies~Object detection}
\ccsdesc[300]{Computing methodologies~Scene understanding}

\keywords{CNN; ball detection; basketball; single viewpoint; low-latency; real-time; dataset; neural networks}

\begin{teaserfigure}
\includegraphics[width=\textwidth]{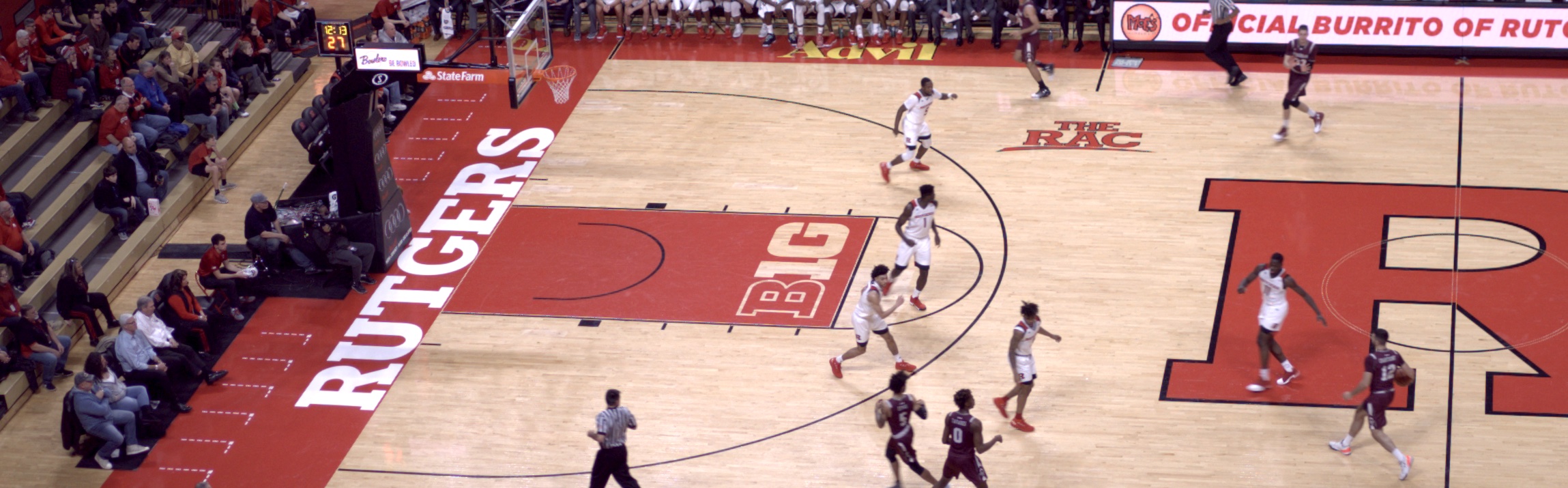}
\caption{Rutgers vs. Fordham at The RAC, 2017 --- Spot the ball.}
\Description{Teaser image of a basketball game}
\label{fig:teaser}
\end{teaserfigure}

\maketitle

\section{Introduction}
\label{sec:intro}

Locating the ball in team sports is an important complement to player detection~\cite{Parisot2017} and tracking~\cite{Kumar2013a}, both to feed sport analytics~\cite{Thomas2017a} and to enrich broadcasted content~\cite{Fernandez2010}.
In the context of real-time automated production of team sports events~\cite{Chen2011,Chen2010d,Chen2010c,Chen2016a,Keemotion}, knowing the ball position with \emph{accuracy} and \emph{without delay} is even more critical.

In the past ten years, the task of detecting the ball has been largely investigated, leading to industrial products like the automated line calling system in tennis or the goal-line technology in soccer~\cite{HawkEye}. However, the ball detection problem remains unsolved for cases of important practical interest because of two main issues.

First, the task becomes quite challenging when the ball is partially occluded due to frequent interactions with players, as often encountered in team sports. Earlier works have generally addressed this problem by considering a calibrated multi-view acquisition setup to increase the candidates detection reliability by checking their consistency across the views~\cite{Parisot2011, Ren2008, intelTrueView, HawkEye, Lampert2012}. To better deal with instants at which the ball is held by players, some works have even proposed to enrich multiview ball detection with cues derived from players tracking~\cite{Wang2014a, Wang2014, Maksai2016}. However, those solutions require the installation of multiple cameras around the field, which is significantly more expensive than single viewpoint acquisition systems~\cite{intelTrueView} and appears difficult to deploy in many venues\footnote{From our personal experience of 100+ installations in professional basketball arenas.}.

Second, the weakly contrasted appearance of the ball, as generally encountered in indoor environments, leads to many false detections. This problem is often handled by exploiting a ballistic trajectory prior in order to discriminate between true and false detected candidates~\cite{Chen2008, Chen2012, Chakraborty2013, Kumar2011, Zhou2013, Yan2006, Parisot2011}.
In addition to this prior, some works also use the cues derived from the players tracking~\cite{Zhang2008, Maksai2016}, or even use the identity and team affiliation of every player in every frame~\cite{Wei2016}.
However, this prior induces a significant delay, making those solutions inappropriate for real-time applications. Furthermore, they require detecting the ball at high frame rates, inducing hardware and computational constraints and making them unusable for single instants applications.

Overall, for scenarios considering instant images in indoor scenes captured from a single view point, the ball detection problem remains largely unsolved. This is especially true for basketball (involving fast dynamics and many players interactions) where state-of-the-art solutions saturates around $40\%$ detection rate, restricted in the basket area~\cite{Parisot2019b}.
Those limitations force current real-time game analysis solutions to rely on a connected ball~\cite{youtube}, which requires to insert a transmitter within the ball ; or use multiple cameras and multiple servers to process the different data streams~\cite{HawkEye, Pingali2000, intelTrueView}.

\smallskip

In this paper, we propose a learning-based method reaching close to $70\%$ ball detection rate on unseen venues, with very few false positives, demonstrated on basketball images. In order to train and validate our method, we created a new dataset of single view basketball scenes featuring many interactions between the ball and players, and low-contrasted backgrounds.

\smallskip

Similar to most modern image analysis solutions, our method builds on a Convolutional Neural Network (CNN)~\cite{Lecun2015a}.
In earlier works, several CNNs have been designed to address the object detection problem, and among the most accurate, \emph{Mask~R-CNN}~\cite{He2017} has been trained to detect a large variety of objects, including balls.
The experiments presented in Section~\ref{subsec:compare} reveal that applying the universal \emph{Mask~R-CNN} detector to our problem largely fails. However, fine tuning the pre-trained weights to deal with our dataset significantly improves the detection performance (despite staying $\sim 5\%$ below our method). However, \emph{Mask~R-CNN} is too complex to run in real-time on an affordable architecture.

To provide a computationally simple alternative to \emph{Mask~R-CNN}, a few works have designed a CNN model to specifically detect the ball in a sports context.
\cite{Reno2018a} adopts a \emph{classification} strategy by splitting the image in a grid of overlapping patches, each of which is fed to a CNN that assesses whether or not it contains a ball. However, the model is far from real-time\footnote{We measured $3.6$fps on an Nvidia~RTX~2080~Ti with an overlap of 10~pixels between the patches. See their paper for implementation details.}, and only two different games are considered to train and validate the method.
In the Robocup Soccer context, \cite{Speck2017a} formulates the detection problem as a \emph{regression} task aiming at predicting the coordinates of the ball in the image. The performance remains relatively poor despite the reasonable contrast
between the white ball and the green field. Furthermore, regression of object coordinates is known to be poorly addressed by CNNs~\cite{Liu2018}. Hence, this strategy is expected to poorly generalize to real team sport scenes.

In comparison to those initial attempts to exploit CNNs to detect the ball, the contributions of our work are multiple and multifaceted.

Primarily, we propose to formulate the ball detection problem as a \emph{segmentation} problem, for which CNNs are known to be quite effective~\cite{Chen2017a}. Such formulation is especially relevant in our team sport ball detection context since there is only one object-of-interest that we aim to detect. Hence, the CNN does not need to handle the object instantiation problem.
In addition, we use a pair of consecutive images coming from a fixed (or motion compensated) viewpoint to take advantage of the ball dynamics without the delay caused by a temporal regularization, allowing low-latency applications.
We show that this approach allows fast and reliable ball detection in weakly contrasted or cluttered scenes.

Furthermore, we show that the use of test-time data augmentation\footnote{Generating multiple transformed versions of an input sample, in order to predict an ensemble of outputs for that input sample} permits a significant increase in the detection accuracy at small false positive rates. This is in the continuation of recent works showing that test-time data augmentation can be used to estimate the prediction uncertainty of a CNN model~\cite{Ayhan2018, Wang2019}.

Finally, we make the royalty-free part of our dataset publicly available\footnote{\url{https://sites.uclouvain.be/ispgroup/Softwares/DeepSport}}.
Beyond providing pairs of consecutive images, it offers a representative sample of professional basketball images gathered on multiple different arenas. It features a large variety of game actions and lighting conditions, cluttered scenes and complex backgrounds. This solves a weakness of several previous works suffering from a limited set of validation data, having very few different games considered~\cite{Parisot2019b, Reno2018a, Speck2017a}.

\section{Method}
\label{sec:model}

\begin{figure*}
    \begin{center}
    \includegraphics[width=0.9\textwidth]{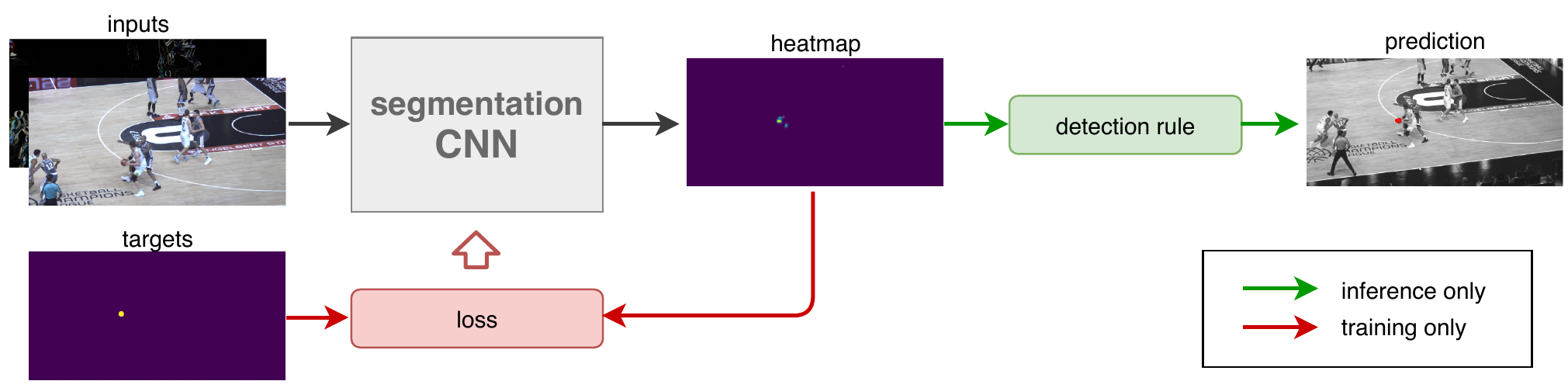}
    \end{center}
    \vspace{-1em}
    \Description{The input image with input difference is given to a lightweight CNN. The heatmap it outputs is compared to the target segmentation mask. At inference, a decision rule is used to infer the ball position from the output heatmap.}
    \caption{
    Our detector is based on a segmentation task performed by a fully convolutional network that outputs a heatmap of the ball position. At inference, a detection rule is used to predict the ball location from the heatmap.}
    \label{fig:overview}
\end{figure*}

Our proposed solution is illustrated in Figure~\ref{fig:overview}. It formulates the ball detection problem as a segmentation problem
where a fully convolutional neural network is trained to output a heatmap predicting the ball segmentation mask.
At inference, a detection rule is used to extract a ball candidate from the heatmap.
The fully convolutional workflow makes it possible to work with any input sizes.

\paragraph{CNN implementation.}
Among fully convolutional neural networks, multi-scale architectures are known to better trade-off complexity and accuracy on large images by combining wide shallow branches that manipulate fine image details through a reduced number of layers, with deep narrow branches that access smaller images and can thus afford a deeper sequence of layers~\cite{Yu2018,Mazzini2019,Poudel2019,Zhao2018}. The major benefit of this tradeoff is a fast processing allowing real-time applications.

In practice, the CNN used to evaluate our method is the ICNet implementation available at~\cite{hellochick} except that (i) the 3 input resolutions were changed from {\footnotesize $\{(1),(1/2),(1/4)\}\times\text{input size}$} to {\footnotesize $\{(1),(1),(1/2)\}\times\text{input size}$} in order to better handle the small size of the ball at the lowest resolutions; and (ii) the input layers were adapted to handle the 6 channels input data (see Eq.~(\ref{eq:diff})). The output heatmap is produced by applying a softmax on the last layer and removing the channel relative to the background.

From a computational point of view, the selected CNN offers satisfying real-time performances (see Section~\ref{subsec:compare} Table~\ref{tab:fps}) but any segmentation network could be used instead; and we could expect even higher detection speed using recent fast segmentation networks~\cite{Yu2018,Mazzini2019,Poudel2019}.

\paragraph{Detection rule.}
We infer ball candidates as points lying above a threshold $\tau$ in the predicted heatmap. Since we know {\it a priori} there is only one ball in the scene, the detection rule is further constrained by limiting the number of ball candidates per scene. This approach, named top-$k$, detects (up to) the $k$ highest spots in the heatmap, as long as they are higher than the threshold $\tau$. In practice, to avoid multiple detections for the same heatmap hotspot, highest points in the heatmap are selected first, and their surrounding pixels are ignored for subsequent detections, in a greedy way.
Note that the computational cost associated with this detection rule is negligible compared to the CNN inference.

\paragraph{Exploiting the ball dynamics.}
When not held by a player, the ball generally moves rapidly, either due to a pass between players, a shot, or dribbles. To provide the opportunity to exploit motion information, we propose to feed the network with the information carried by two consecutive images denoted $\mathcal{I}_a$ and $\mathcal{I}_b$, where each image is composed of the 3 conventional channels in the $RGB$ space.
Two strategies were considered. The first one, consisting in feeding the network with the concatenation of $\mathcal{I}_a$ and $\mathcal{I}_b$ on the channels axis, gave poor results and is not presented in this work. The second strategy consists in concatenating the image of interest together with its difference with the previous image. Hence:
\begin{equation}
    \small
  \text{Input} =
  \left(R_{\mathcal{I}_a},G_{\mathcal{I}_a},B_{\mathcal{I}_a},
  \lvert R_{\mathcal{I}_a}-R_{\mathcal{I}_b}\rvert ,
  \lvert G_{\mathcal{I}_a}-G_{\mathcal{I}_b}\rvert,
  \lvert B_{\mathcal{I}_a}-B_{\mathcal{I}_b}\rvert \right)
\label{eq:diff}
\end{equation}

\paragraph{Training.}
Because of the custom number of input channels required by our method, no pre-training of the network is done.
The training is performed using the Stochastic Gradient Descent optimization algorithm applied on the mean cross-entropy loss, at the pixel level, between the output heatmap and the binary segmentation mask of the ball.
The meta parameters were selected based on grid searches. The learning rate has been set to $0.001$ (decay by a factor of $2$ every $40$ epochs), batch size to $4$, and number of epochs to $150$ in all our experiments. The weights obtained at the iteration with the smallest error on the validation set were kept for testing.
The network is fed with $1024\times512$ pixels inputs obtained by a data augmentation process including mirroring (around vertical axis), up- and down- scaling (maintaining the ball size between 15 and 45 pixels, which corresponds to the size range of balls observed by the cameras used to acquire the dataset) and cropping (keeping the ball within the crop).

\paragraph{Terminology.}
We will use the term \emph{\bf scene} when referring to the original image captured by the camera, and \emph{\bf random-crop} to denote a particular instance of random cropping, scaling, and mirroring parameters. Multiple different random crops can thus be extracted from a single scene.

\section{Validation Methodology}
\label{sec:validation}

\begin{figure*}
    \begin{center}
    \includegraphics[width=\textwidth]{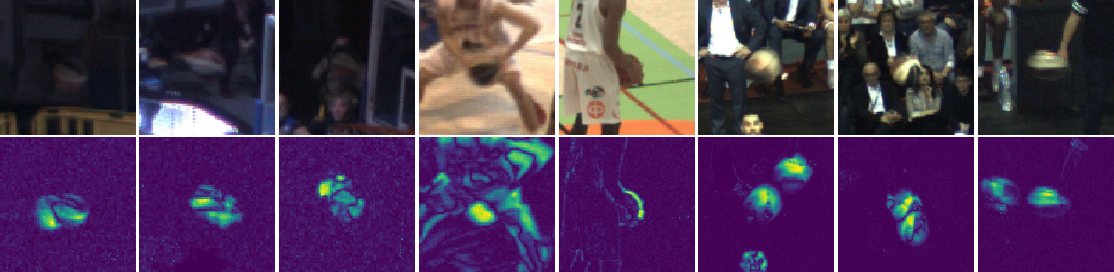}
    \end{center}
    \vspace{-0.5em}
    \caption{Image samples (top) and corresponding mean differences to previous image (bottom) for scenes in which the ball is detected when providing the difference to the previous image to the network, but remains undetected by the model that ignores the difference.}
    \Description{difference between two consecutive images shows the circular shape of the ball}
    \label{fig:where_diff_helps}
\end{figure*}

\paragraph{Dataset.}
\label{sec:dataset}
The experiments were conducted on a rich dataset of 280 basketball scenes coming from professional games that occurred in 30 different arenas on multiple continents. The cameras captured half of a basketball court and have a resolution between 2Mpx and 5Mpx. The resulting images have a definition varying between 65px/m (furthest point on court in the arena with the lowest resolution cameras) and 265px/m (closest point on court in the arena with the highest resolution cameras). For each scene, two consecutive images were captured. The delay between those two captures is 33ms or 40ms, depending on the acquisition frame rate. As shown in Figure~\ref{fig:results}, the dataset presents a large variety of game configurations and various lighting conditions. The ball was manually annotated in the form of a binary segmentation mask. In all images, at least half of the ball is visible.

\paragraph{K-fold testing.}
To validate our work, a $K$-fold training/testing strategy has been adopted in all our experiments. It partitions the dataset into $K$ subsets, named folds, and runs $K$ iterations of the training/testing procedure. Each iteration preserves one fold for testing, and shuffle the other folds before splitting the samples in $90\%$ for the training set and $10\%$ for the validation set.
To assess the generalization capabilities of our model to unseen games and arenas, the $K$ folds were defined so that each fold only contains images from arenas that are not present in the other folds.
In practice, $K$ has been set to 7, which means that each fold contains about 40 scenes coming from 3 or 4 different arenas.

\paragraph{ROC curves and metrics.}
For a given top-$k$ detection rule, the accuracy is assessed based on ROC curves. Each ROC plots the detection rate (TPR) as a function of the false positive rate (FPR), while progressively changing the detection threshold $\tau$. The detection rate measures the fraction of scenes for which the ball has been detected, while the false positive rate measures the mean number of false candidates that are detected per scene. In practice, a detection is considered as being a true (false) ball detection if it lies inside (outside) the surface covered by the ball in the annotated segmentation mask.

\paragraph{Test-time data augmentation.}
Multiple different random crops from the same scene were considered at test time by aggregating their output heatmap. As the ball location is unknown at inference, the random crops that were combined were similar (IoU $> 0.9$). It is like providing small variations of an input in which the ball is expected to be detected (see Figure~\ref{fig:stacking_strategy}~(left)).

\section{Experimental results}
\label{sec:results}

This section assesses our Ball Segmentation method (named \emph{BallSeg}) based on the dataset introduced in Section~\ref{sec:validation}.
First, we validate the CNN input and the detection rule.
Then, we compare \emph{BallSeg} with the \emph{Mask~R-CNN} fine tuned with our dataset.
Finally, we use test-time data augmentation to analyze how different random crops impact the detection. In addition, we take advantage of test-time data augmentation to improve detection accuracy at the cost of increased computational complexity, by merging the heatmaps associated with multiple random crops of the same scene.

\subsection{Inputs choice and detection rule validation}

\begin{figure}
    \begin{center}
    \includegraphics[width=0.45\columnwidth]{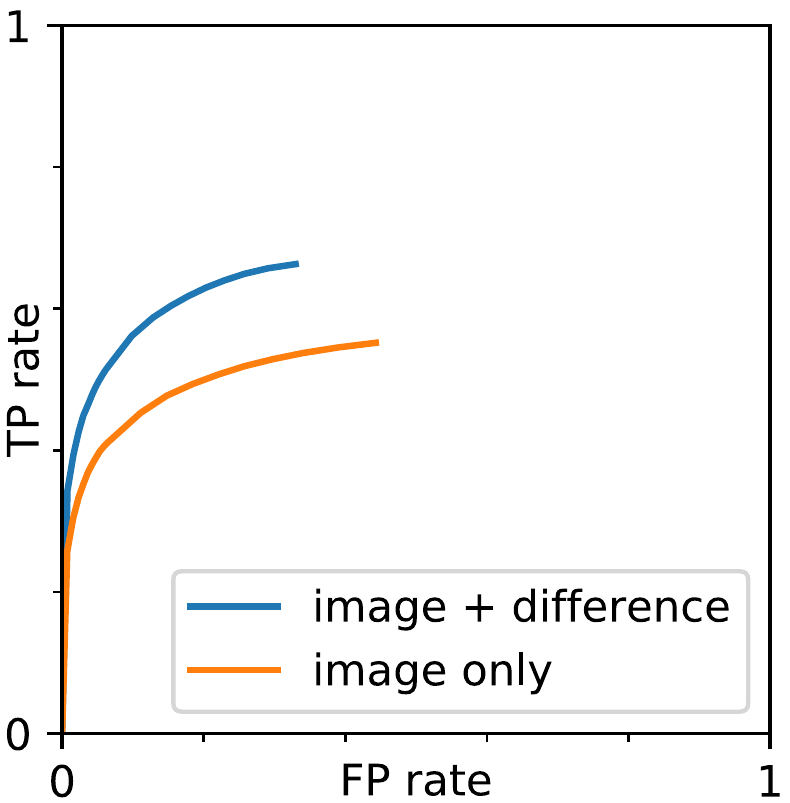}\hspace{2em}
    \includegraphics[width=0.45\columnwidth]{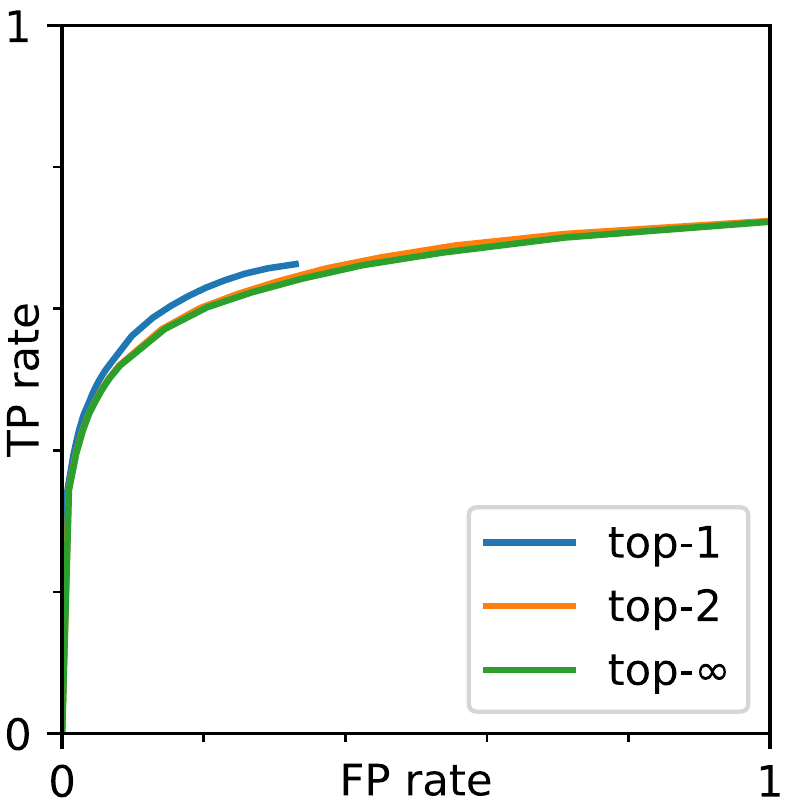}
    \end{center}
    \caption{\emph{BallSeg} ROC curves. Left: adding the difference with the previous image to the network input improves accuracy (top-$1$ detection rule). Right: top-$1$ detection rule achieves better true/false positives trade-offs.}
    \Description{ROC curves for image + difference and image only}
    \label{fig:roc_baseline}
\end{figure}

We compared the performances of our \emph{BallSeg} when feeding the network with and without the difference between consecutive images. Figure~\ref{fig:roc_baseline} (left) shows that presenting the input as described in Eq.~(\ref{eq:diff}) significantly improves the detection accuracy. Figure~\ref{fig:where_diff_helps} reveals that using the difference with the previous image allows detecting the ball in low contrasted scenes.

Different top-$k$ detection rules are compared in Figure~\ref{fig:roc_baseline} (right). A top-$1$ detection rule achieves better true/false positive trade-offs at low false positive rates (for $\tau=0.01$: {\sc tpr}=$0.66$ and {\sc fpr}=$0.33$). This is far better than all previous methods presented in the introduction, especially given that our dataset includes a large variety of scenes, presenting the ball in all kinds of game situations.

In the rest of the paper, unless specified otherwise, results use a top-$1$ detection rule, and the input described by Eq.~(\ref{eq:diff}).

\begin{figure*}
    \begin{center}
    \includegraphics[width=\textwidth, trim={28pt, 11pt, 21pt, 5pt}, clip]{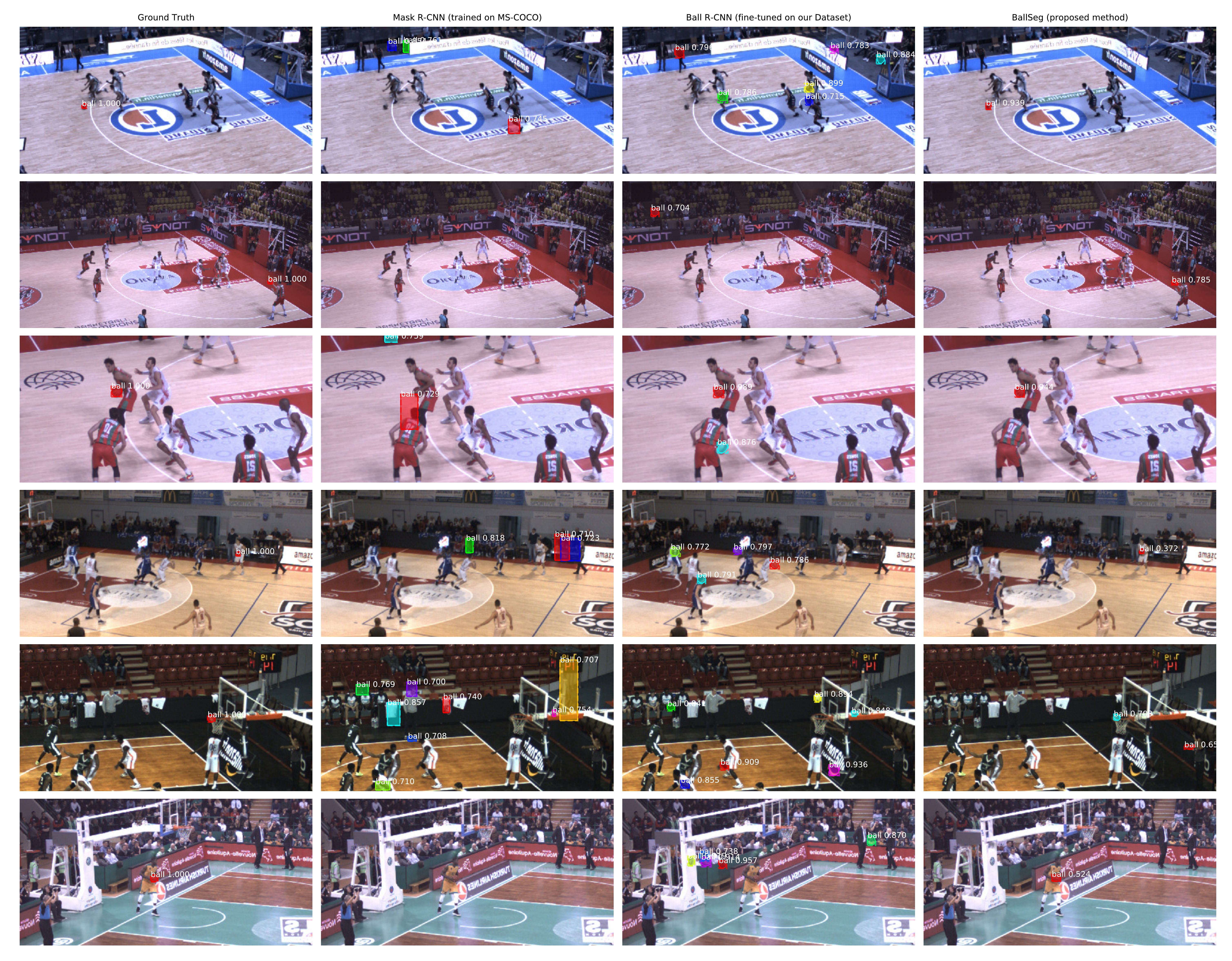}
    \hrule
    \includegraphics[width=\textwidth, trim={28pt, 10pt, 21pt, 1pt}, clip]{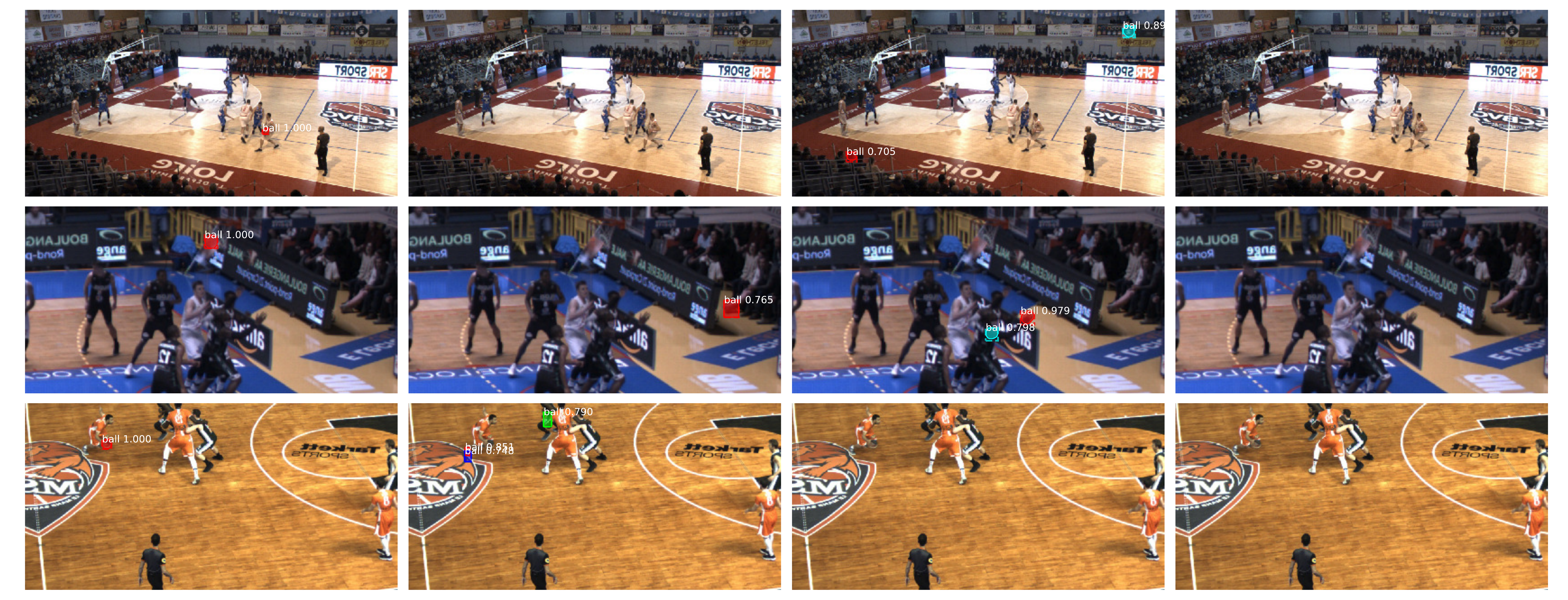}
    \end{center}
    \vspace{-10pt}
    \caption{Our \emph{BallSeg} gives accurate results in many different configurations. Top 6 rows: success cases. Bottom 3 rows: failure cases. First column: \emph{Ground-Truth}. Second column: the class "ball" from the out-of-the-shelf \emph{Mask~R-CNN}. Third column: fined-tuned \emph{Mask~R-CNN} on our dataset (\emph{Ball~R-CNN}). Fourth column: our method, using a segmentation CNN (\emph{BallSeg}).}
    \label{fig:results}
    \Description{success cases and failure cases}
\end{figure*}

\subsection{Comparison with state-of-the-art object detection method}
\label{subsec:compare}

To better evaluate the value of our \emph{BallSeg} model, we compared it with the results obtained with an universal \emph{Mask~R-CNN} model. We used the implementation of \cite{matterport_maskrcnn_2017} that provides weights trained on the MS-COCO dataset~\cite{coco} and already has a class for the ball.
For a fairest comparison, the \emph{Mask~R-CNN} network has also been fine tuned to deal with our dataset. The resulting model (denoted \emph{Ball R-CNN}) was obtained by running a conventional optimization with stochastic gradient descent optimizer with a learning rate $lr$. First, we updated the front part of the network for $n_f$ epochs, then we updated the whole network for $n_w$ epochs. A grid search has been performed to select the $(lr, n_f,n_w)$ triplet. Best $K$-fold mean test set accuracy was obtained with $lr=10^{-3}$, $n_f=10$ and $n_w=20$.

\begin{align*}
    lr &\in \{10^{-2}, 10^{-3}, 10^{-4}, 10^{-5}\} \\
    n_f &\in \{0, 1, 10\} \\
    n_w &\in \{0, 1, 10, 20, 100\}
\end{align*}

\begin{figure}
    \begin{center}
    \includegraphics[width=0.45\columnwidth]{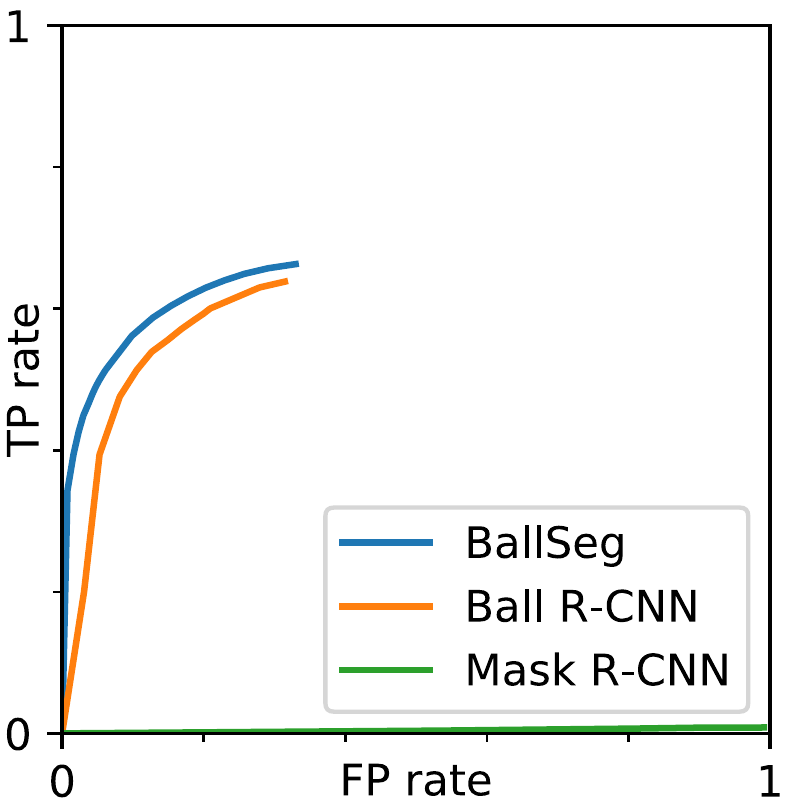}\hspace{2em}
    \includegraphics[width=0.45\columnwidth]{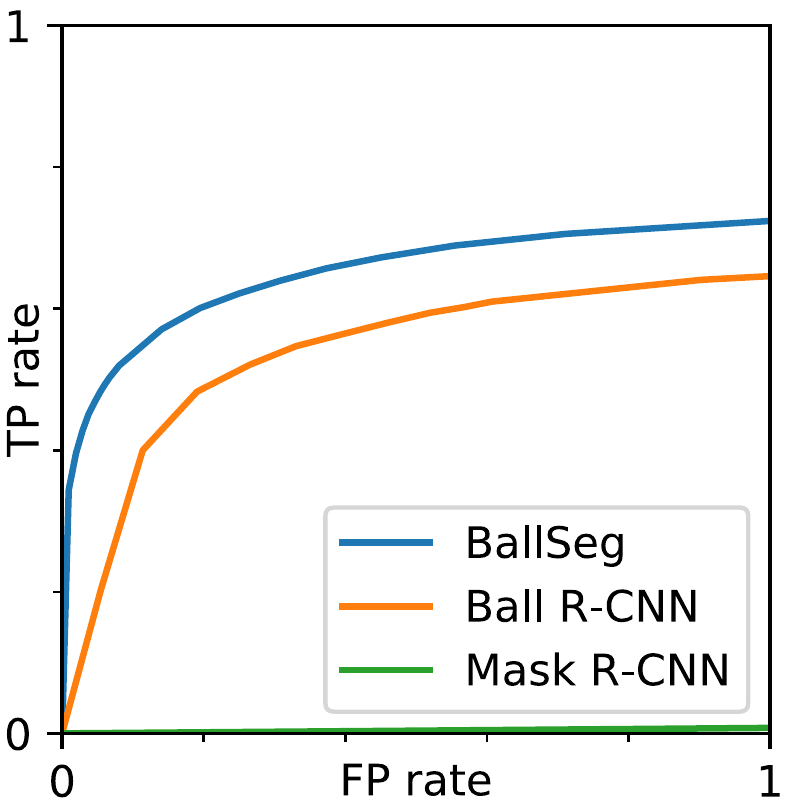}
    \end{center}
    \caption{Comparison between \emph{BallSeg} and two computationally complex \emph{Mask~R-CNN} instances: \emph{Mask~R-CNN} pre-trained on MS-COCO, and the same network fine tuned on our dataset (\emph{Ball~R-CNN}). Left: top-$1$ detection rule; Right: top-$2$ detection rule.}
    \label{fig:roc_rcnn}
    \Description{ROC curves with Mask-R-CNN against BallSeg}
\end{figure}

In Figure~\ref{fig:roc_rcnn}, we observe that \emph{BallSeg} outperforms \emph{Ball~R-CNN}. Besides, the fact that \emph{Mask~R-CNN} largely fails reveals the need to have sport-specific datasets for the complex task of ball detection in a team sport context. Figure~\ref{fig:results} shows a visual comparison between \emph{BallSeg}, the out-of-the-shelf \emph{Mask~R-CNN} and \emph{Ball~R-CNN} on different image samples from our dataset.

In addition to this qualitative analysis, Table~\ref{tab:fps} presents the measured computational complexity of the two models: \emph{BallSeg} using the \emph{ICNet} implementation from~\cite{hellochick} and \emph{Ball~R-CNN} being \emph{Mask~R-CNN} from~\cite{matterport_maskrcnn_2017}. 
We observe that, besides reaching a better accuracy than \emph{Ball~R-CNN}, \emph{BallSeg} is significantly faster. Note that those numbers highly depends on the implementation. Indeed, \emph{ICNet} can benefit from a $6\times$ speed improvement once compressed with filter-pruning, while keeping the same accuracy~\cite{icnet_implem,fss,Li2019}.

\newcommand{\shape}[3]{{\footnotesize $ #1 \times #2 \times #3 $}}
\begin{table}
    \begin{tabular*}{\columnwidth}{l@{}rrr}
    \toprule
                                &     single image     &     image + diff     &     image + diff     \\
                                & \shape{1024}{512}{3} & \shape{1024}{512}{6} & \shape{1280}{720}{6} \\
    \midrule
    \emph{Mask~R-CNN}           &       4.33 fps      &          N/A          &          N/A         \\
    \emph{BallSeg} (our method) &      38.39 fps      &       24.67 fps       &       12.08 fps      \\
    \bottomrule
    \end{tabular*}
    \caption{Framerate of the two methods compared on an Nvidia~GTX~1080~Ti, with a batch size of $2$, without using filter-pruning optimization. The segmentation approach is significantly faster than the state-of-the-art \emph{Mask~R-CNN}.}
    \label{tab:fps}
\end{table}

\subsection{Test-time data augmentation}
\label{subsec:testtimeaugment}

This section investigates how different random crops of the same scene impacts the detection. It first observes
that the ball segmentation errors rarely affect all random crops. In other words, prediction errors correspond to high entropy output distributions, which by definition correspond to a high uncertainty. This is in line with the observation made by~\cite{Ayhan2018, Wang2019} that wrong predictions are associated with high uncertainty levels, i.e. to large diversity of predictions for transformed inputs.
More interestingly, our experiments also reveal that false positives are not spatially consistent across the ensemble of heatmaps. This is an important novel observation, since it gives the opportunity to significantly increase ($\sim10\%$) the detection accuracy at small false positive rate by aggregating the ensemble of heatmaps obtained by multiple different random-crops.

\subsubsection{Random crops consistency and failure case analysis}

\begin{figure}
    \begin{center}
    \includegraphics[width=\columnwidth]{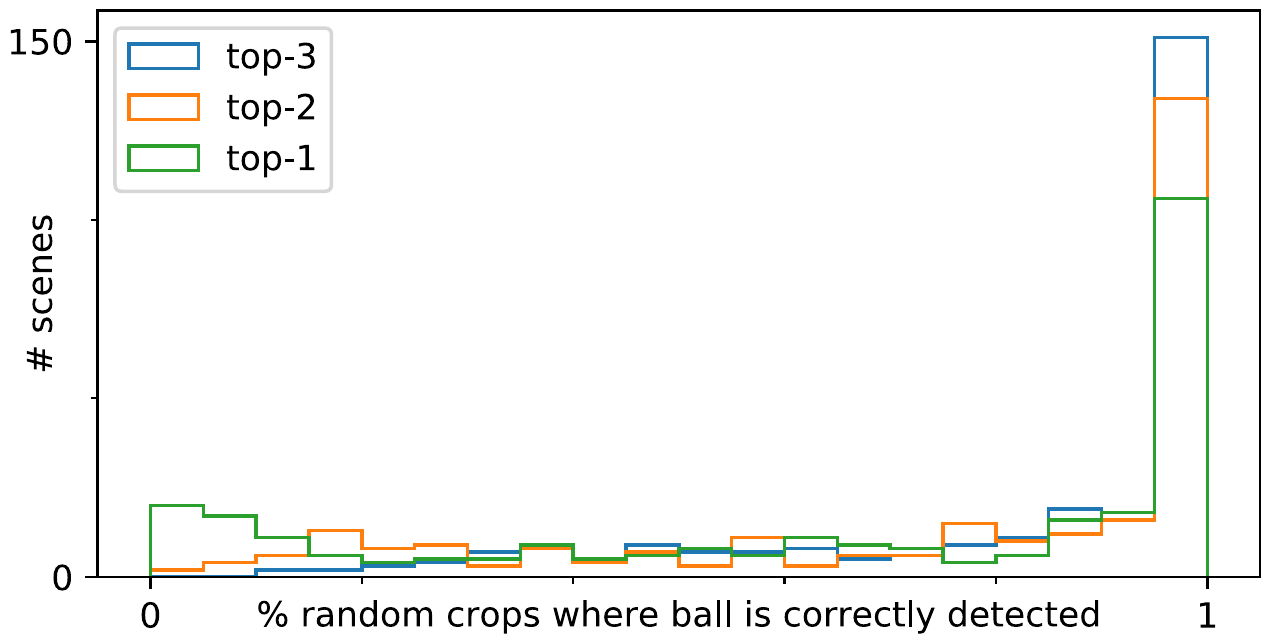}
    \caption{1000 random-crops were created for each scene.
    Distribution, over the 280 dataset scenes, of the percentage of random crops in which the ball is detected for top-$1$, top-$2$, and top-$3$ detection rules.}
    \label{fig:topk_helps}
    \end{center}
    \Description{multiple random crops are considered}
\end{figure}

\begin{figure}
    \begin{center}
    \includegraphics[width=\columnwidth]{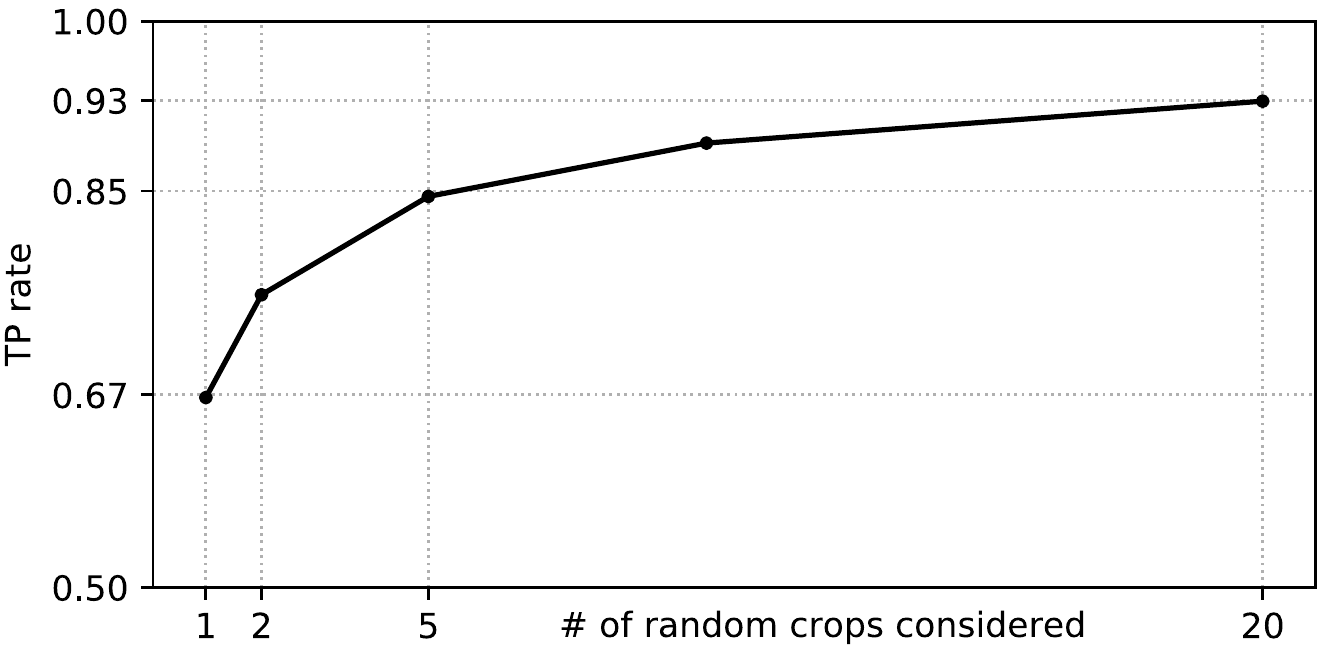}
    \caption{1000 random-crops were created for each scene.
    Ball detection rate as a function of the number of random crops considered for a scene, for a top-$1$ detection rule. The ball is detected in the scene if it is detected in at least one of the random crops considered.}
    \label{fig:multiple_random_crops}
    \end{center}
    \Description{multiple random crops are considered}
\end{figure}

\begin{figure*}
    \begin{center}
    \includegraphics[width=0.33 \textwidth, trim={30pt, 10pt, 20pt, 10pt}, clip]{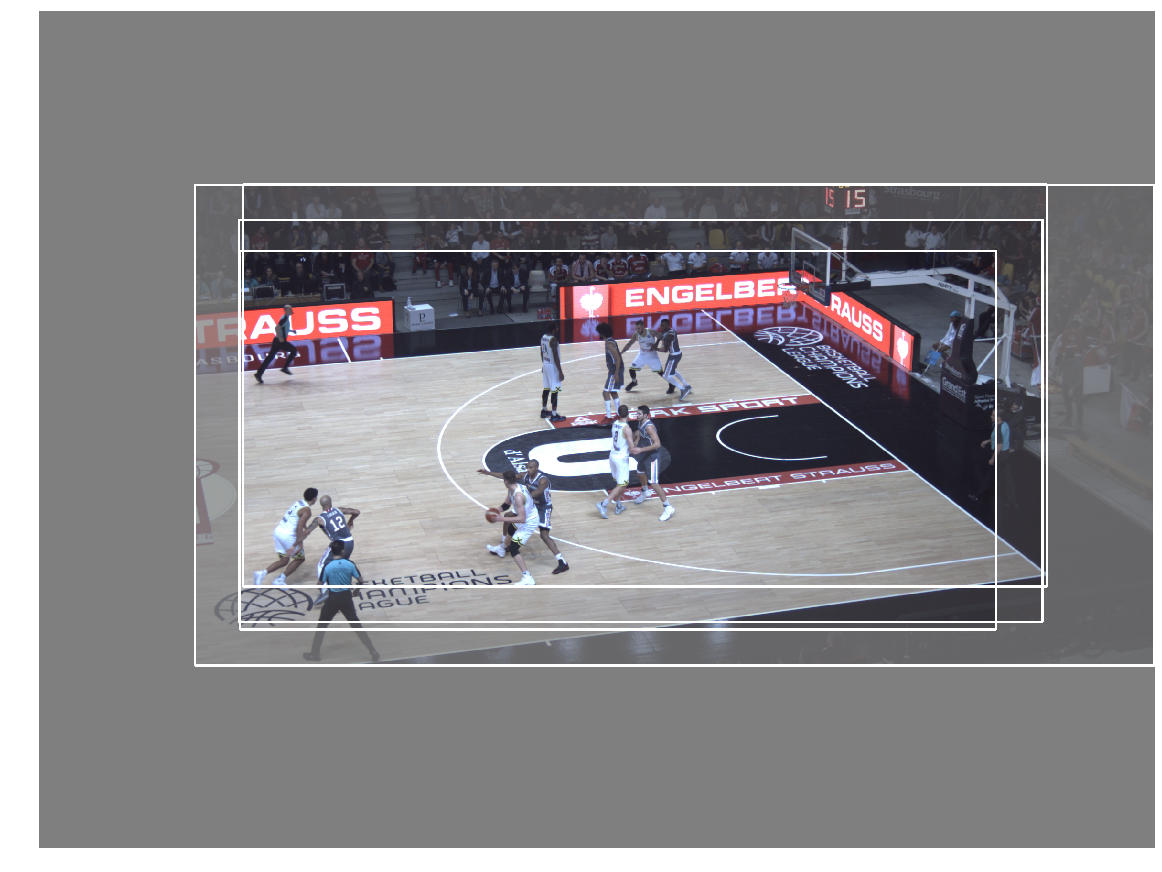}
    \includegraphics[width=0.33 \textwidth, trim={30pt, 10pt, 20pt, 10pt}, clip]{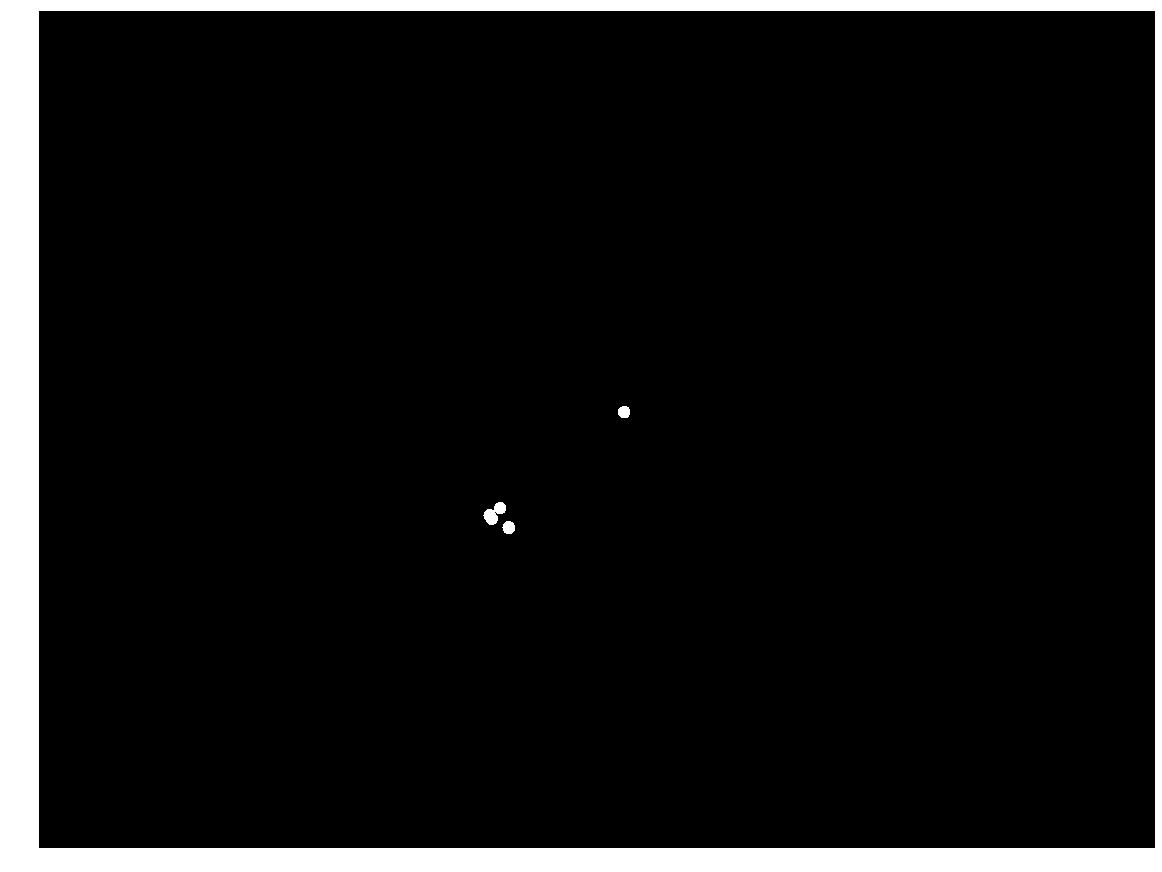}
    \includegraphics[width=0.33 \textwidth, trim={30pt, 10pt, 20pt, 10pt}, clip]{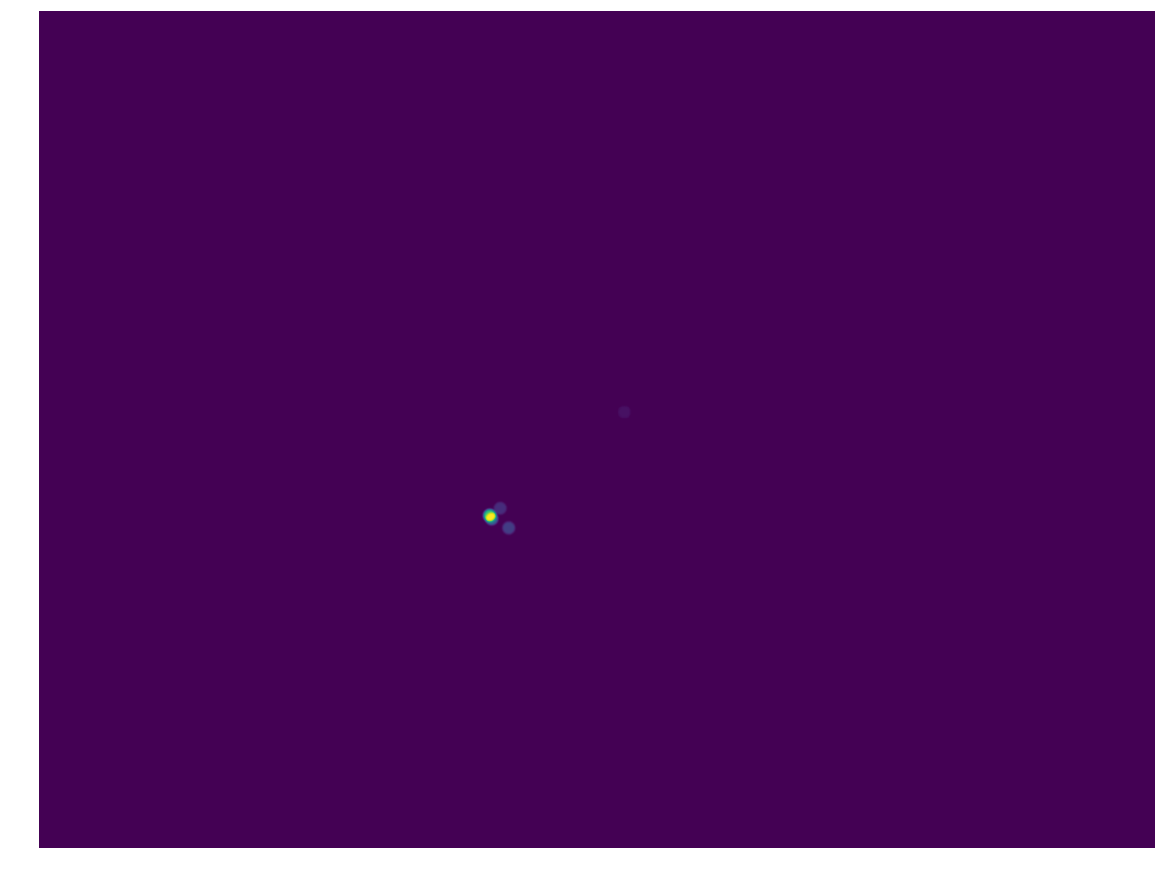}
    \end{center}
    \vspace{-10pt}
    \caption{Averaging the heatmaps over random image samples helps to discriminate the ball among candidates. Left: Five random crops having similar IoU in a given scene are used; Middle: Scatter of their heatmaps most salient points; Right: Mean heatmap intensity using the five input heatmaps.}
    \Description{The agregation of multiple heatmaps}
    \label{fig:stacking_strategy}
\end{figure*}

Figure~\ref{fig:topk_helps} presents the distribution, over our 280 scenes, of the percentage of random crops in which the ball is detected for top-$1$, top-$2$, and top-$3$ detection rules and $\tau=0$.
We observe that when the ball is not detected in all random crops (about half of the scenes), it is generally detected in some random crops.
This suggests that the detection rate could be improved by combining multiple random crops (see Section~\ref{subsec:tradeoff}). This opportunity is supported by Figure~\ref{fig:multiple_random_crops}, which presents, for a top-$1$ detection rule, the ball detection rate as a function of the number of random crops considered for a scene. We observe that for $93\%$ of the dataset scenes, when 20 different random crops are considered for each scene, the ball is detected as top-$1$ in at least one of those random crops.

\subsubsection{Accuracy/complexity trade-off}
\label{subsec:tradeoff}

This section investigates how to improve the true/false positive trade-off based on the computation of more than one heatmap per scene.
It builds on the fact that the ball has a higher chance to be detected if more random crops are considered (see Figure~\ref{fig:multiple_random_crops}),
and on the observation that the ball generally induces a significant spot in the heatmap of every random crop, even if this spot is not always the highest one (see in Figure~\ref{fig:topk_helps} where a top-$2$ or top-$3$ detection rule allows to increase the number of scenes in which the ball is detected for all random crops).

\begin{figure}
    \begin{center}
    \includegraphics[width=0.45\columnwidth]{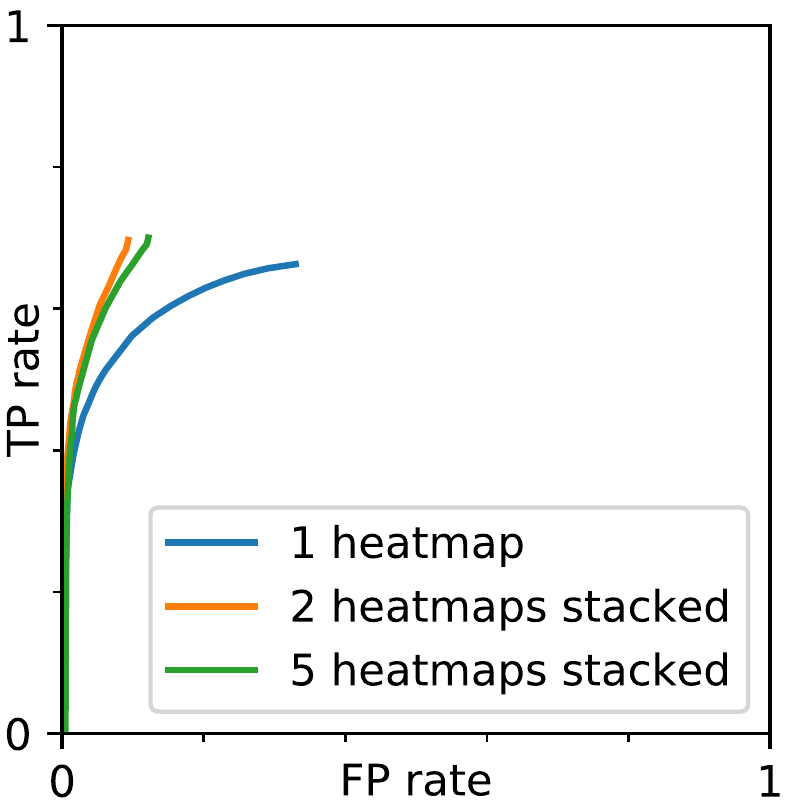}\hspace{2em}
    \includegraphics[width=0.45\columnwidth]{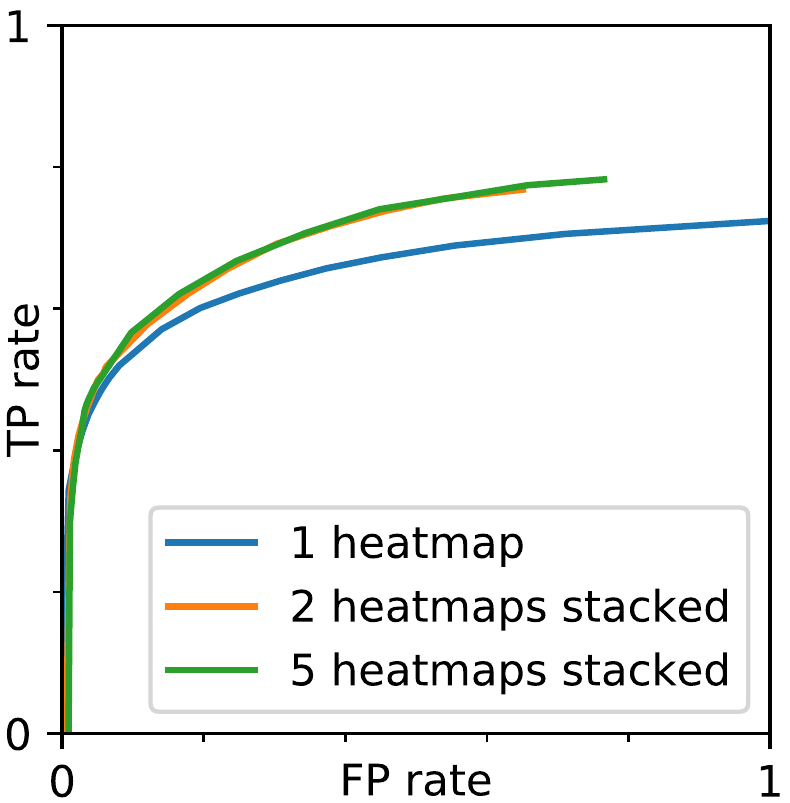}
    \caption{Left: top-$1$ and Right: top-$2$ ROC curves obtained when heatmaps coming from $1$, $2$ or $5$ similar random crops of the same scene are averaged.}
    \label{fig:roc_multiple_images}
    \end{center}
    \Description{ROC multiple images}
\end{figure}

Figure~\ref{fig:stacking_strategy} shows an example where the heatmaps predicted for distinct random crops of the same scene are generally more consistent at the actual ball location than for the false candidates.
From this observation, we propose to merge the heatmaps of different random image samples by averaging their intensity. Figure~\ref{fig:roc_multiple_images} shows that applying a top-$1$ detection rule on the averaged heatmap significantly improves the true vs. false positive rate tradeoff ($\sim 10\%$ detection rate increase at small false positive rate), already when only two random image samples are averaged.

This gives the opportunity to improve accuracy at the cost of additional complexity, which in turn can support accurate automatic annotation of unlabelled data for more accurate models training, thereby following the data distillation paradigm~\cite{Radosavovic2018}.

\section{Conclusion}
\label{sec:conclusion}

This paper proposes a new approach for ball detection using a simple camera setup by (i) adopting a CNN-based segmentation paradigm, taking advantage of the ball uniqueness in the scene, and (ii) using two consecutive frames to give cues about the ball motion while keeping a very low latency.

Furthermore, the approach benefits from recent advances made by fast segmentation networks, allowing real-time inference.
In particular, the segmentation network used to demonstrate the approach reduces drastically the computational complexity compared to the conventional \emph{Mask~R-CNN} detector, while achieving better results: close to $70\%$ detection rate (with a very small false positive rate) on unseen games and arenas, based on an arduous dataset.

The dataset made available with this paper is unique in term of number of different arenas considered. In addition, we show that, despite being relatively small, it offers enough variety to provides good performances, especially for a method that doesn't use pre-trained weights.

\section*{Acknowledgments}
This research is supported by the DeepSport project of the Walloon Region, Belgium. C. De Vleeschouwer is funded by the F.R.S.-FNRS (Belgium).
The dataset was acquired using the Keemotion automated sports production system.
We would like to thank Keemotion for participating in this research and letting us use their system for raw image acquisition during live productions, and the LNB for providing rights on their images.
\begin{center}
\begin{tabular}{cc}
\raisebox{0.12em}{\includegraphics[width=2cm]{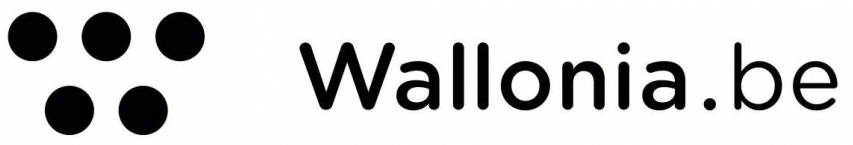}} & \includegraphics[width=1.8cm]{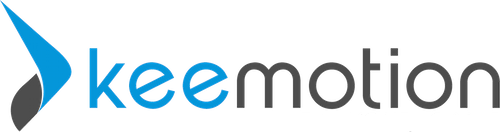}
\end{tabular}
\end{center}

\bibliographystyle{ACM-Reference-Format}
\balance
\bibliography{ballseg}

\end{document}